\documentclass[preprint]{elsarticle}

\usepackage[utf8]{inputenc}
\usepackage{multirow}
\usepackage{graphicx}
\usepackage{indentfirst}
\usepackage[small]{caption}
\usepackage{multirow}
\usepackage{booktabs}
\usepackage{algorithm}
\usepackage{algorithmic}
\usepackage{bm}

\usepackage{amsfonts}
\usepackage{amssymb}
\usepackage{amsthm,amsmath}
\usepackage{mathrsfs}
\usepackage{indentfirst}
\usepackage{multirow}
\begin{document}

\begin{frontmatter}

\title{Zero-sample surface defect detection and classification based on semantic feedback neural network}

\author{Yibo~Guo, Yiming~Fan, Zhiyang~Xiang, Haidi~Wang, Wenhua~Meng~and~Mingliang~Xu}

\begin{abstract}
Defect detection and classification technology has changed from traditional artificial visual inspection to current intelligent automated inspection, but most of the current defect detection methods are training related detection models based on a data-driven approach, taking into account the difficulty of collecting some sample data in the industrial field. We apply zero-shot learning technology to the industrial field. Aiming at the problem of the existing "Latent Feature Guide Attribute Attention" (LFGAA) zero-shot image classification network, the output latent attributes and artificially defined attributes are different in the semantic space, which leads to the problem of model performance degradation, proposed an LGFAA network based on semantic feedback, and improved model performance by constructing semantic embedded modules and feedback mechanisms. At the same time, for the common domain shift problem in zero-shot learning, based on the idea of co-training algorithm using the difference information between different views of data to learn from each other, we propose an Ensemble Co-training algorithm, which adaptively reduces the prediction error in image tag embedding from multiple angles. Various experiments conducted on the zero-shot dataset and the cylinder liner dataset in the industrial field provide competitive results.
\end{abstract}


\end{frontmatter}

\section{Introduction}

Defect detection and classification have important research significance in different research and application fields. Many defect detection technologies have been specially developed for specific application fields\cite{Dirac1953888,Feynman1963118,sun2019effective}. Also in the production or use of industrial products, there are often different surface defects, such as wear, crack, nicks, etc. This situation is generally because the physical and chemical properties of a certain area of the product surface are inconsistent with the surrounding background, This will affect the aesthetics of the product and cause a serious decline in product quality\cite{he2020deep}, so a lot of work has been done in the detection and classification of surface defects\cite{czimmermann2020visual,kumar2008computer,le2020novel}.

Traditional defect detection and classification are often artificial visual inspections. Quality inspection workers need to use related magnifying instruments to detect product surface defects. This has problems such as high labor costs, low efficiency, and low accuracy. Automated inspection methods based on machine vision can alleviate the above problems, so they have gradually become a trend to replace artificial visual inspection. Early defect detection methods based on machine vision used more artificially guided low-level visual features combined with traditional machine learning methods, by extracting global features of the image or by SIFT\cite{torralba2003context}, HOG\cite{dalal2005histograms}, SURF\cite{bay2008speeded}, LBP\cite{ojala1996comparative} and other methods extract the local features of the image, then use neural network\cite{dong2019pga}, SVM\cite{shumin2011adaboost} and other machine learning methods to achieve defect detection. Although traditional methods have achieved good results in defect detection and classification problems in the industrial field, this method needs to be based on rich professional knowledge and artificial fine-tuning. Deep learning abandons this tedious and inefficient artificial design. The feature step is to learn the inherent feature distribution of data based on a data-driven approach by designing a multilayer neural network. This deep model obtained through learning has achieved relatively advanced performance in image analysis and recognition. Wei et al. proposed a four-stage defect detection framework\cite{he2020deep}, and formulating the defect detection problem into pixel segmentation and defect classification based on regression, which not only can provides pixel-level defect severity and defect types, but also makes users more flexible in use. Song L et al. improved the U-Net framework and proposed DU-Net\cite{song2019weak} based on the Deep Convolutional Neural Network (DCNN), using two white light sources as the lighting system, and deconvolution instead of upsampling so that the model can extract more features. At the same time, the number of feature channels is doubled in the down-sampling step, so that the network input and output sizes are the same to avoid the edge loss of the image. The author proves the effectiveness of this method through experiments.

Although the performance of related detection models has surpassed the human level\cite{xie2017aggregated}, training these deep models requires a large amount of labeled data, and these models can only identify the target classes included in the training set, and lack the ability to identify target classes that are not included in the training set\cite{palatucci2009zero}.The setting of fully supervised learning is unrealistic in most industrial problems. Through communication with companies, we learned that it is difficult to collect a large number of defective samples during product production or use, and even if the collection is completed, the relevant models need to be retrained so that the entire process is costly. At the same time, the product may produce defects that have not occurred in the past.
Recently, Lampert et al.\cite{lampert2009learning} proposed a Zero-shot Learning(ZSL) method that is very different from traditional image classification tasks. They used different types of animal images in the training phase and the test phase, and proposed DAP and IAP using the semantic attributes shared by the training set (also called unseen classes) and the test set (also called seen class) as a bridge, where the semantic attribute represents the feature of a target instance or a certain latent attribute, and it can be a specific text description, or it can be a feature vector that does not have semantic interpretation. The method of DAP uses the image features of the training set to learn a classifier for each attribute, thereby learning the relationship between image features and attributes, and in the test stage, the classifier is used to predict the attributes of other animals, and then the new animals in the test set are judged by comparing with the attributes in the attribute library. The method of IAP is to first estimate the probability of whether the picture belongs to the known class, and then use the relationship between the attributes contained in the seen class and the unseen class to predict the test sample.

ZSL mainly studies how to use the common attribute information of the training set and the test set to classify the test set when the data categories are mutually exclusive and there is no intersection, it can overcome a series of problems caused by missing samples or difficulty in collecting, which also provides ideas for other fields when facing sample problems\cite{gao2019know, rahman2020improved,rezaei2020zero}. Zhu et al. proposed a zero-shot target detection architecture ZS-YOLO\cite{zhu2019zero}, which provides a semantic description of the target and corrects the confidence loss and part of the backbone network in the YOLOV2 framework to the rules required by the ZSL task, thus realizing the detection of unseen targets, and proved the rationality and effectiveness of its method through analysis and experiments. In view of the difficulty and cost of data collection in industrial faults, Feng et al. proposes an attribute transfer method based on fault descriptions through the ZSL idea \cite{feng2020fault}, the detection of unknown faults is realized by describing the causes and conditions of each fault, and using these descriptions as auxiliary knowledge sources.

\begin{figure}[t]
\centering
\includegraphics[scale=0.3]{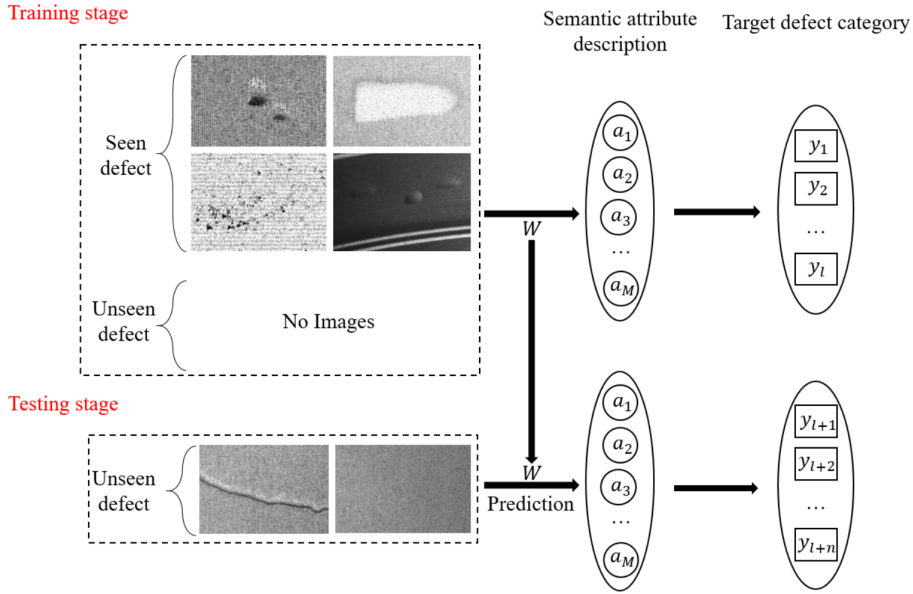}
\caption{Surface defect detection and classification process}
\label{fig:1}
\end{figure}

In response to the above problems and the needs of future development in the defect detection field, we have studied related defect detection scenarios that do not provide training samples of the target defect category, instead of conventional defect detection and classification. Figure.\ref{fig:1} shows the basic process of the zero-sample industrial surface defect detection method, in the training phase, a large number of marked seen defects are used, but there are no unmarked unseen defects, we can identify unseen defects through the semantic attribute description shared by the two. Attributes act as a bridge between the training class and the test class, and make up for the asymmetry between the two in terms of information, making it feasible to complete zero-sample defect detection and classification tasks. In terms of attributes, we consider from the aspects of versatility and accuracy, the word vector extracted by the Glove \cite{pennington2014glove} is used as an intermediate medium between the image and the category label, shared between the unseen defect and the seen defect, and propose a "Latent Feature Guide Attribute Attention" (LFGAA) network based on semantic feedback to realize the detection effect of unseen defects, thereby replacing the conventional defect detection and classification tasks.

The main contributions of this paper are as follows:

\begin{itemize}
\item We tried defect detection and classification without the target sample, and used the word vector extracted by the Glove model as the auxiliary knowledge source to realize the task of zero-sample defect detection and classification.
\item Pointed out the latent problems of different distribution of LGFAA networks in space, and proposed an LGFAA network based on semantic feedback to alleviate this problem.
\item For the common domain shift problem in ZSL, an Ensemble Co-training (ECT) algorithm for zero-sample classification is proposed, and its effectiveness is shown in the representative dataset.
\end{itemize}

The rest of this paper is arranged as follows: In the second part, we propose the problem of zero-sample industrial defect detection and classification, then introduce the selection process of semantic attributes and the analysis process of the LGFAA network. In the third part, the details of our proposed SF-LFGAA network and ECT algorithm are introduced. After that, the various modules of the proposed method are analyzed to prove the rationality of the proposed method. Finally, the paper is summarized.

\section{Preliminaries}
\subsection{Zero-sample defect detection and classification problem definition}
The focus of Zero-sample defect detection and classification problem is to complete the knowledge transfer from seen classes to unseen classes, aiming to identify images without the defects in the training stage for detection and classification, that is, to detect n kinds of unknown situations defined by the test set.

In this problem, the defective sample of the training set is defined as $D^{S}$, The defective sample label is $Y^{S}=\left\{y_{1}^{s}, \ldots, y_{l}^{s}\right\}, D^{S}=\left\{x_{1}^{s}, \ldots, x_{l}^{s}\right\}$ has $l$ pictures that have been tagged and there are $m$ kinds of defects, $x_{i}^{s}$ represents the i-th picture in $D^{S}, y_{i}^{s}$ is the category label of $x_{i}^{s}$ and $y_{i}^{s} \in Y^{S}$. The defect samples of the test set are defined as $D^{U}$ and label as $Y^{U}=\left\{y_{l+1}^{u}, \ldots, y_{l+q}^{u}\right\}$, $D^{U}=\left\{x_{l+1}^{u}, \ldots, x_{l+q}^{u}\right\}$ total of $q$ unmarked pictures and $n$ kinds of defects, Where $y_{i}^{u} \in Y^{U}$. The training set and the test set have no intersection in categories, that is, $Y^{U} \cap Y^{S}=\emptyset$ and $Y^{U} \cup Y^{S}=Y$. For every $y \in Y$, there is a semantic attribute $A_{y}=\left\{a_{1}, \ldots, a_{l+t}\right\}$ related to it, where t represents the semantic attribute dimension of the defect.
\subsection{Semantic attributes of cylinder liner dataset}

\begin{figure}[t]
\centering
\includegraphics[scale=0.3]{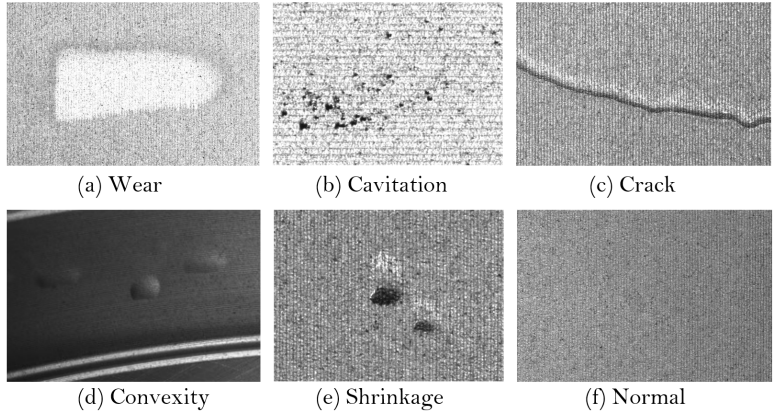}
\caption{CLSDD Dataset}
\label{fig:2}
\end{figure}

Semantic attributes represent the characteristics or latent attributes of a target instance. Part of the feasibility of zero-shot learning depends on whether the middle-level semantic attributes have sufficient discrimination and expressiveness \cite{xian2018zero}, due to the lack of semantic attribute annotations in industrial datasets, so we summarized the semantic attribute description of each defect based on the relevant literature\cite{you2018influence,shastin2020laser,huang1999composites} and the description of some company employees. Take the cylinder liner data set CLSDD used in the following experiments as an example, as shown in Figure.\ref{fig:2}, CLSDD has 6 categories, including 5 defect categories (cavitation, wear, crack, convexity, shrinkage). Intuitively, when there are defects on the surface of our image, generally speaking, we first notice its defect characteristics and want to understand the related causes, so the description of the defect is composed of the defect color, range, shape, and reason for the formation of the defect, and the related attributes shown in Table \ref{tab:1} are formulated.

\begin{table*}[htb]
\centering
\begin{center}
\caption{CLSDD dataset defect data attributes} \label{tab:1}
\begin{tabular}{cc}
\hline
\toprule
Attribute number&Define attributes \\
\hline
1 & Defect color: white \\
2 & Defect color: black \\
3 & Defect color: gray \\
4 & Defect scope area: large \\
5 & Defect scope area: medium \\
6 & Defect scope area: small \\
7 & Defect shape: irregular long strip \\
8 & Defect shape: Irregular ellipse \\
9 & Defect shape: irregular dense shape \\
10 & Defect reason: related to liquid metal parameters \\
11 & Defect reason: related to cooling water \\
12 & Defect reason: related to engine \\
13 & Defect reason: related to the body shaft bearing \\
14 & Defect reason: related to three filters \\
15 & Defect reason: related to lubricating oil \\
\bottomrule
\end{tabular}
\end{center}
\end{table*}

Limited by the lack of professional knowledge on the causes of cylinder liner defects, this binary vector did not achieve the desired effect in the later experiments. At the same time, we consider that this situation even works better in some applications\cite{feng2020fault}, the labeling of related attributes is not a simple labeling category label, that is, expert-level artificial labeling is required, which is difficult to promote in most cases in other fields. Because we use the GloVE model to extract word vectors, the GloVE model is trained through open source corpora such as Wikipedia. The detected image category can be input into it to obtain a multi-dimensional vector representation. This vector can better contain the semantic relationship between words and instead of artificially labeling attributes. We use it as an intermediate medium between the image and the category label, and share it between the unknown defect and the known defect, so as to achieve the detection effect of the test category. This kind of semantic feature may be less accurate than the expert-level artificially defined attributes in the zero-shot dataset, but this is currently the most used attribute labeling method for no attribute annotation data. The CLSDD dataset uses the GloVe model to extract the word vector label names respectively "cylinder liner crack, cylinder liner wear, cylinder liner convexity, cylinder liner shrinkage, cylinder liner cavitation, cylinder liner normal", and finally selected a 200-dimensional GloVe vector for the experiment.

\subsection{LFGAA network}

\begin{figure}[t]
\centering
\includegraphics[scale=0.45]{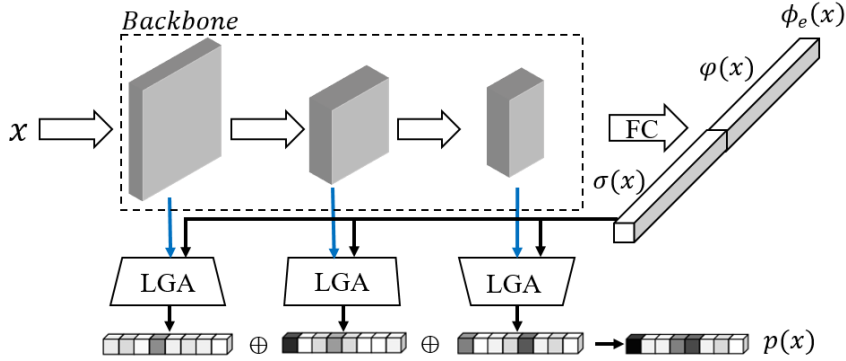}
\caption{Latent Feature Guide Attribute Attention (LFGAA) Network}
\label{fig:3}
\end{figure}

Our method is based on the LFGAA network\cite{liu2019attribute}, which has the leading effect in the zero-shot image classification network. It uses the latent attributes proposed in the LDF\cite{li2018discriminative} to increase the discrimination between categories, and proposed the object-based attribute attention $p(x)$ to solve the problem of equal treatment of semantic attributes in the conventional end-to-end zero-sample classification network, which led to the model's misclassification in some categories. As shown in Figure.\ref{fig:3}, the visual features of the image output by Backbone are output as a 2K-dimensional augmented space $\phi_{e}(x)$ through the fully connected layer, where 2K-dimensional is composed of the artificially defined attribute space (K-dimension) and the latent attribute space (K-dimension). The artificially defined attribute prediction $\varphi(x)$ and the latent attribute prediction $\sigma(x)$ are obtained respectively:

\begin{equation}
\phi_{e}(x)=[\varphi(x) ; \sigma(x)], \quad \varphi(x), \sigma(x) \in \mathbb{R}^{k}
\end{equation}
For the latent attributes $\sigma(x)$, the goal is to make the learned features distinguishable for object recognition. Here we use Triplet loss to learn latent discriminant attributes $\sigma(x)$ by adjusting the distance between categories:

\begin{equation}
\mathcal{L}_{\text {lat }}=\max \left(0, m+d\left(\sigma\left(x_{i}\right), \sigma\left(x_{k}\right)\right)-d\left(\sigma\left(x_{i}\right), \sigma\left(x_{j}\right)\right)\right)
\end{equation}

Where $x_{i}$  and $x_{k}$ are images from the same category, and $x_{i}$ and $x_{j}$ are images from different categories. $d(x, y)$ is the Euclidean distance between $x$ and $y$, and $m$ is the margin of the triple loss. And $\phi(x)$ uses attributes as labels and optimizes through softmax loss function. In the prediction phase, in order to perform zero-shot image classification in the latent attribute space, it is necessary to know the latent attribute prototype of the unseen class. Unlike artificially defined attributes, the latent attributes cannot be directly obtained. It is necessary to obtain the latent attribute features of all seen classes first. According to the category to average, the formula is as follows:

\begin{equation}
\overline{\sigma^{S}}=\frac{1}{N} \sum_{x \in x_{i}^{s}} \sigma\left(x_{i}\right)
\end{equation}

Where $\overline{\sigma^{S}}$  represents the prototype of the latent attributes of the seen class, and then calculates the correlation between the seen class and the unseen class in the artificially defined attributes by means of ridge regression:

\begin{equation}
\beta_{c}^{u}=\operatorname{argmin}\left\|A_{y}^{u}-\sum_{c \in Y^{s}} \beta_{c}^{u} A_{y}^{c}\right\|_{2}^{2}+\lambda\left\|\beta_{c}^{u}\right\|_{2}^{2}
\end{equation}

Where $A_{y}^{s}$ is the semantic attribute of the seen class, $\quad A_{y}^{u}$ is the semantic attribute of the unknown class, $\beta_{c}^{u}$ is the mapping parameter between the artificially defined attributes to be learned, and then this correlation is transferred to the latent attribute prototype of the seen class, so as to obtain the latent attribute characteristics of the unseen class:

\begin{equation}
\overline{\sigma^{u}}=\sum_{c \in Y^{s}} \beta_{c}^{u} \overline{\sigma^{s}}
\end{equation}

Finally, the category of the unseen class is obtained in the following way:

\begin{equation}
y^{*}=\operatorname{argmax} S\left(\sigma(x), \overline{\sigma^{u}}\right)
\end{equation}

Which $S(x, y)$ refers to the similarity between the predicted semantic attribute and the unseen semantic attribute, which can be expressed by cosine similarity or inner product. The predicted category of the unseen class is judged by comparing the predicted attribute with the semantic attribute of each unseen class.

\begin{figure}[t]
\centering
\includegraphics[scale=0.35]{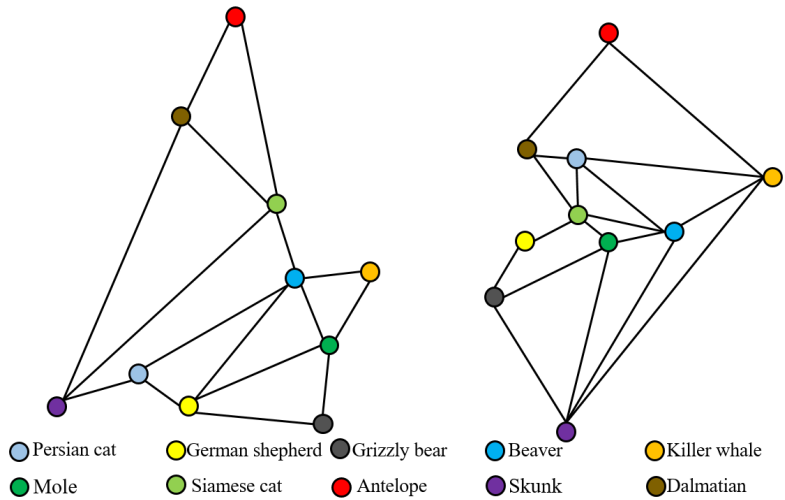}
\caption{The visual distribution of latent attribute prototypes (left) and artificially defined attributes (right)}
\label{fig:4}
\end{figure}

Formula (3-5) transfers the semantic attribute correlation between artificially defined attributes to the latent attribute space. This is because the model’s prediction of the unseen class needs to be compared with the latent attribute prototype of the unseen class constructed through attribute correlation. However, the spatial distribution of the artificially defined attributes and the learned latent attributes may be different. Figure.\ref{fig:4} shows the latent attribute prototypes and real semantic attributes visualized by T-SNE for some categories in the AwA2 data set, it can be seen from this that the spatial distribution of the learned latent attribute prototypes and real semantic attributes is obviously different, this may cause the accuracy of the model to decrease.

\section{Method}

\begin{figure}[t]
\centering
\includegraphics[scale=0.39]{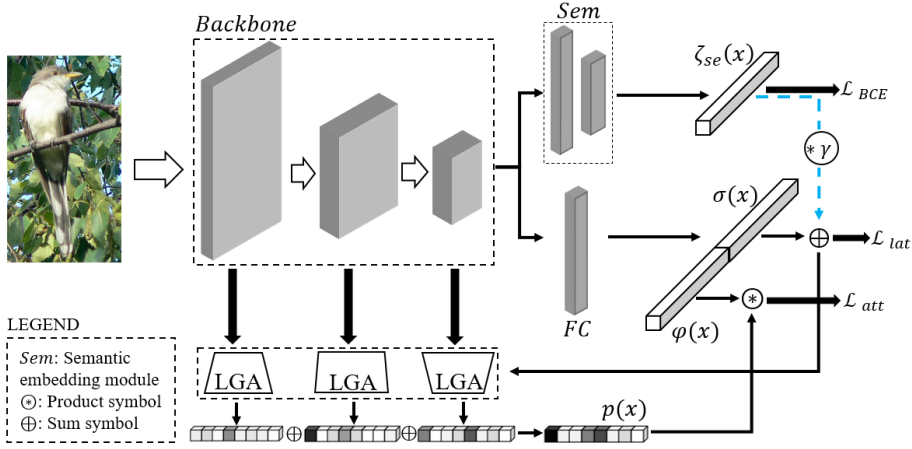}
\caption{SF-LFGAA network architecture}
\label{fig:5}
\end{figure}

According to the different description of the spatial distribution of latent attributes and artificially defined attributes, we propose a semantic embedding module and feedback mechanism based on the LGFAA network to alleviate this situation. The overall network is shown in Figure.\ref{fig:5}. Specifically, the network is divided into two parts and share the Backbone. One part is the original part of the LGFAA network, which combines Backbone with the fully connected layer to learn the mapping from visual space to semantic space and latent attribute space to obtain predicted semantic attributes, latent attributes and target-based attribute attention; the other part is the combination of Backbone and the semantic embedding module to learn the mapping from the visual space to the semantic space, and then the feedback mechanism feeds back the semantic vector output by the semantic embedding module to the latent attribute space to obtain the modified latent attribute. For unseen class predictions, SF-LFGAA and LFGAA networks use the same prediction method.

We use artificially defined attributes as the semantic embedding module of attribute labels to join the network for training, and using the feedback mechanism, the semantic vector output by the semantic embedding module is used to adjust the latent attributes of the LGFAA network output, which to reduce the gap with artificially defined attributes, the latent attributes constructed through attribute correlation are more discriminative, and the acquisition of attribute attention also requires the combination of visual features and latent attributes, which can further reduce the ambiguity between target categories.

\subsection{Semantic Embedded Module}

In the LFGAA network, there is also a semantic space besides the latent attribute space. The reason why we rebuild a semantic embedding module is because the attribute attention mechanism makes non-representative semantic attributes not highly active, and uses it as semantic embedding for adjustment the latent attributes may prevent the model performance from being significantly improved, and at the same time impose additional constraints on the network.

The semantic embedding module consists of two fully connected layers and a ReLU activation function. The semantic embedding module maps the deep visual features of the image to the semantic space. Unlike most existing zero-sample classification methods based on embedding\cite{akata2015evaluation}, we do not directly use the depth features of the pre-trained CNN model as its visual representation. It combines Backbone with semantic embedded modules to optimize the network in an end-to-end manner. Specifically, the image features output from Backbone are input to several fully connected layers, and finally activated by ReLU. In order to ensure that the semantic embedding of the output is close to the artificially defined attributes, the binary cross entropy loss function is used to Semantic embedding module is optimized:

\begin{equation}
\mathcal{L}_{B C E}=-\frac{1}{N} \sum_{i=1}^{N}\left(a_{i} \log \tau\left(\zeta_{s e}\left(\phi\left(x_{i}\right)\right)\right)+\left(1-a_{i}\right)\left(1-\log \tau\left(\zeta_{s e}\left(\phi\left(x_{i}\right)\right)\right)\right)\right)
\end{equation}

Where $\zeta_{s e}(x)$ is the semantic vector output by the semantic embedding module, $\phi(x)$ is the visual feature vector output by Backbone, and $\tau$ represents the sigmoid activation function.

Since we share Backbone with the LGFAA network, we need to train the two parts at the same time to optimize the network. The triple loss function used by the LGFAA network has been introduced in Section 2.3. The training latent attribute space and the semantic space use the Triple loss function and Softmax loss function:
\begin{equation}
\mathcal{L}_{\text {lat }}=\frac{1}{N} \sum_{i}^{N}\left[\left\|\sigma\left(x_{i}\right)-\sigma\left(x_{j}\right)\right\|^{2}-\left\|\sigma\left(x_{i}\right)-\sigma\left(x_{k}\right)\right\|^{2}+\alpha\right]_{+}
\end{equation}
\begin{equation}
\mathcal{L}_{a t t}=-\frac{1}{N} \sum_{i}^{N} \log \frac{\exp \left(\varphi\left(x_{i}\right)^{T} \operatorname{diag}\left(p\left(x_{i}\right)\right) a_{y_{i}}\right)}{\sum_{y \in Y^{s}} \exp \left(\varphi\left(x_{i}\right)^{T} \operatorname{diag}\left(p\left(x_{i}\right)\right) a_{y}\right)}
\end{equation}

Where $\sigma(\mathrm{x})$ represents the latent attributes output by the LFGAA network, $x_{i}, x_{j}$ are images from the same category, $x_{i}, x_{k}$ are images from different categories, $\alpha$ is the margin of the triple loss,$\varphi(x)$ is the output semantic vector that artificially defines the attribute space(ie, semantic space), $p(x)$ is the attribute attention, and $diag$ represents the diagonal matrix function.

Finally, combining these three optimization objectives with the balance factors $\beta_{1}$ and $\beta_{2}$ is the final loss function used to train the SF-LFGAA network:
\begin{equation}
\mathcal{L}=\mathcal{L}_{l a t}+\beta_{1} \mathcal{L}_{a t t}+\beta_{2} \mathcal{L}_{B C E}
\end{equation}

\subsection{Feedback mechanism}

In the process of constructing the network, we considered that there may be a certain gap between the latent attributes output by the LGFAA network and the artificially defined attributes. Therefore, the correlation between the artificially defined attributes may not be suitable for the latent attributes. The semantics are embedded in the module through the feedback mechanism. The output is used to adjust the latent attributes, making the latent attributes more discriminative.

\begin{figure}[t]
\centering
\includegraphics[scale=0.47]{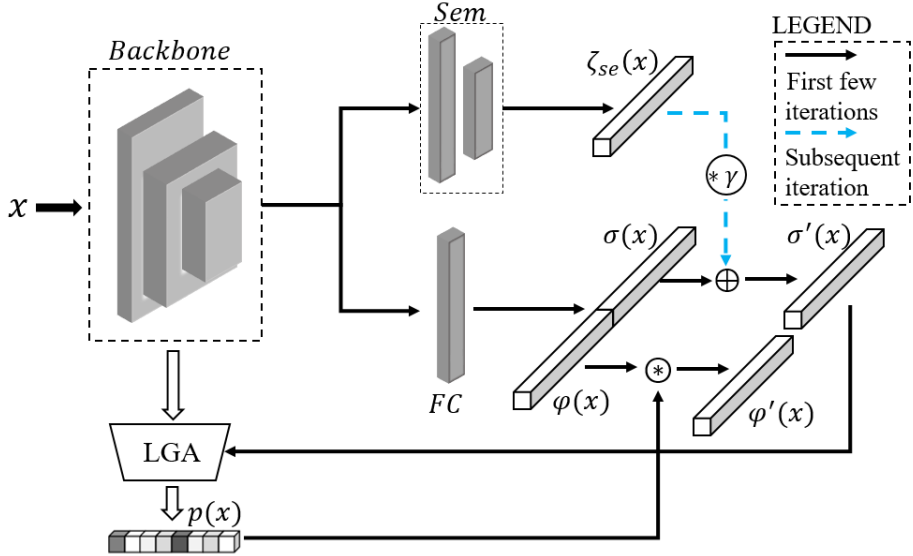}
\caption{Feedback mechanism}
\label{fig:6}
\end{figure}

The feedback mechanism iteratively refines the latent attributes in both the training phase and the unseen class prediction phase, as shown in Figure.\ref{fig:6}. This feedback mechanism effectively uses the semantic embedding module in both the network training and the unseen class prediction phase, and the output of the semantic embedding module is used as the input of the feedback mechanism to adjust the latent attributes. The new latent attribute acquisition formula is as follows:

\begin{equation}
\mathcal{L}=\mathcal{L}_{\text {lat }}+\beta_{1} \mathcal{L}_{a t t}+\beta_{2} \mathcal{L}_{B C E}
\end{equation}

Where $\gamma$ controls the degree of adjustment of the feedback mechanism

It can be seen from Figure.\ref{fig:6} that when the semantic embedding module is first trained, since the semantic attributes of its output may be too far from the artificially defined attributes, the feedback mechanism is only implemented after a few rounds of network training. In the LFGAA network, the acquisition of attribute attention requires the combination of different depths of visual features in Backbone and latent attributes, and then the combination of attribute attention and predicted semantic attributes as the final output of the network, so in addition to the semantic embedding module, the adjusted latent attributes can also affect the Backbone module of the network, prompting the network to generate more discriminative features, and further reducing the ambiguity between target categories.

\subsection{Zero-shot image classification prediction}

In order to prove the effectiveness of the proposed method, we use the same prediction method as the LGFAA network to prove the effectiveness of the proposed semantic embedding module and feedback mechanism.

It can be seen from the SF-LFGAA network structure that the output of the network has latent attributes and semantic attributes. For the prediction of latent attributes, it has been explained by formula (3-6) when the problem of latent attribute space is introduced in section 2.3.

For the prediction of semantic attributes, since the semantic attribute $\zeta_{s e}(x)$ output by $Sem$ only plays a role in adjusting the latent attributes, only the semantic attribute $\varphi(x)$ output by the part of the LGFAA network is used for prediction, and compare it directly with artificially defined attributes to find the closest category. In this paper, the latent attribute space and the artificially defined attribute space are considered at the same time, and the prediction is continued through the following formula:

\begin{equation}
y^{*}=\operatorname{argmax}\left(S\left(\varphi(x), A_{y^{u}}\right)+S\left(\sigma(x), \overline{\sigma^{u}}\right)\right)
\end{equation}

\subsection{Ensemble Co-training algorithm}

The ultimate goal of ZSL is to predict unseen classes. There are many similar attributes between seen and unseen classes, but they often contain different appearances (such as tails. Pig tails are very different in appearance compared to tigers, zebras and other animals), which may lead to the problem of domain shift\cite{fu2015transductive}, so that when the model recognizes the unseen class, it is often classified into the seen class, resulting in a decrease in model accuracy.

In response to this problem, inspired by the co-training algorithm\cite{blum1998combining} and ensemble learning\cite{dietterich2002ensemble}, we propose an Ensemble Co-training algorithm (ECT) for ZSL. The algorithm uses the difference information between different views of data or different classifiers to learn from each other, and combines unlabeled samples to optimize the performance of classifiers or related network models. Literature\cite{he2019bag,han2018semi} has proved that for the same dataset, the models learned from different network structures have different prediction distributions for the test set. The basic idea of the ECT we proposed is (1) Construct the view features of different network structures of the image to replace the different views required in the co-training algorithm, and ensemble multiple different classifiers to adaptively reduce the prediction error in the image label embedding from multiple angles. (2) In the training phase, reliable unseen data is added to alleviate the domain shift problem.

\begin{figure}[t]
\centering
\includegraphics[scale=0.3]{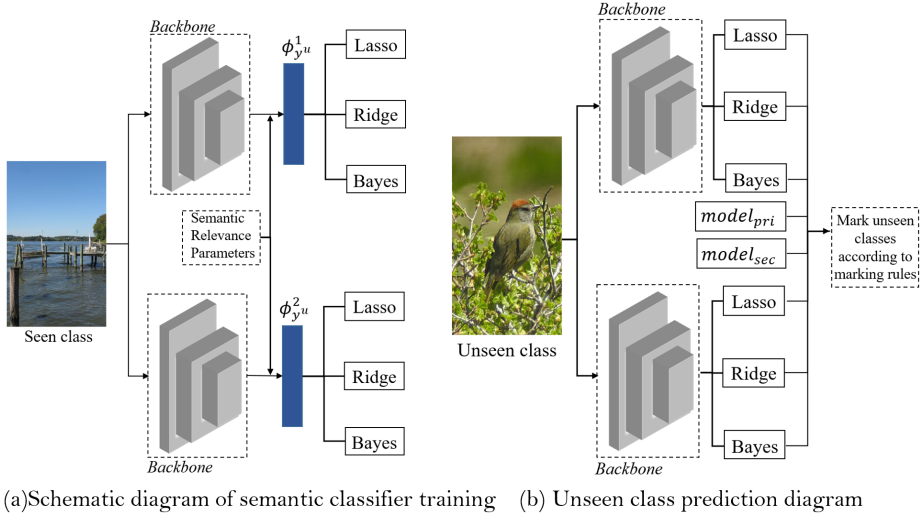}
\caption{Schematic diagram of pseudo-label prediction process}
\label{fig:7}
\end{figure}

In the ECT step, we first use the LFGAA method to train networks with different structures, and get the predicted pseudo-labels, which are called primary and secondary networks based on their performance, and build a collaborative network. After that, the correlation between the attributes of the seen class and the unseen class is transferred to the prototype feature and the feature standard deviation of the seen class image to construct the virtual feature of the unseen class. The prototype feature is the central feature of all samples of each class, and the correlation between attributes $\beta_{c}^{u}$ is obtained by formula(4). Then the visual prototype feature $\phi_{y^{s}}$  of the seen class and the virtual prototype feature $\phi_{y^{u}}$  of the unseen class are obtained in the following way:

\begin{equation}
\phi_{y^{s}}=\frac{1}{N} \sum_{x \in x_{i}^{s}} \phi(x)
\end{equation}
\begin{equation}
\phi_{y^{u}}=\sum_{c \in Y^{s}} \beta_{c}^{u} \phi_{c}
\end{equation}

Similarly, use the above method to obtain the feature standard deviation of the seen class, and calculate the feature standard deviation of the unseen class, and then synthesize a certain number of virtual visual features that conform to the Gaussian distribution for each unseen class $\phi_{y_{i}^{u}}^{i}$, due to the use of Based on the features extracted by different networks, the unknown virtual features synthesized by different networks are respectively $\phi_{y_{i}^{u}}^{i}$ and $\phi_{y_{i}^{u}}^{2}$, and use it as the training data of the attribute classifier in the proposed algorithm. Figure.\ref{fig:7}(a) shows the training process of the semantic classifier. By extracting the visual prototype features and standard deviations of the seen classes, and combining the semantic correlation parameters to construct the visual features of the unseen class, then used to train the semantic classifier. Where the classifier uses Lasso regression, ridge regression, and Bayesian ridge regression. The classifier learns the mapping of visual features to semantic attributes in the following ways:

\begin{equation}
L_{c l a}\left(\phi_{y_{i}^{u}}^{i}, A_{i}^{u} ; W_{c l a}\right)=\phi_{y_{i}^{u}}^{i} W_{c l a} A_{i}^{u}
\end{equation}

Where $W_{\text {cla }}$ represents the parameters of the semantic classifier used, $A_{i}^{u}$ is the semantic attribute of the synthesized unseen class feature $\phi_{y_{i}^{u}}^{i}$. After that, the trained classifier predicts the unseen class, and selects the credible samples for labeling. Figure.\ref{fig:7}(b) shows the process of pseudo-label prediction and labeling, where model $_{\text {pri }}$ and model $_{\text {sec }}$ are pseudo-labels for collaborative network prediction. The prediction results of other classifiers are extracted through the collaborative network to extract unseen class features and input into the classifier for prediction:

\begin{equation}
y_{i}^{u}=\operatorname{argmax} S\left(\phi\left(x_{i}^{u}\right) W_{c l a}, A_{y}^{u}\right)
\end{equation}

After obtaining the two pseudo-labels predicted by the collaborative network itself and the prediction result of the semantic classifier, the unseen class is labeled according to the labeling rules (Step 4) in the algorithm, where N classifier voting rules come from ensemble learning, which means that when no less than N classifiers predict the same category, the sample is considered to be a reliable sample, and the sample is marked as a predicted pseudo label. The detailed steps of the Ensemble Co-training algorithm are as follows:
\begin{algorithm}
\caption{Ensemble Co-training Algorithm}
\label{alg:A}
\hspace*{0.02in}{\bf Input:}
Seen class $D^{S}$, Unseen class $D^{U}$\\
\hspace*{0.02in}{\bf Output:}
SF-LFGAA model\\
\textbf{Initialization:}
\begin{algorithmic}
\STATE $D_{\text {train }} \leftarrow D^{S}, D_{\text {pseudo }} \leftarrow \emptyset$;
\STATE Train two LGFAA networks with different structures on $D_{\text {train }}$ to form a collaborative network, and get the pseudo-labels of the two networks to predict $D^{U}$;
\end{algorithmic}
\textbf{Repeat}
\begin{algorithmic}[1]
\STATE The collaborative network extracts $D^{S}$ and $D^{U}$ features separately, Combine the $\beta_{c}^{u}$ obtained by the formula (4) with the prototype features and feature standard deviations of the known image to synthesize a certain number of virtual features of the unknown class $\phi_{y^{u}}^{1}, \phi_{y^{u}}^{2}$;
\STATE Use $\phi_{y^{u}}^{1}, \phi_{y^{u}}^{2}$ to predict the pseudo-labels of $D^{U}$ features by formula (15-16);
\STATE Choose the best 5 pseudo-labels from the 8 pseudo-labels predicting the $D^{U}$ category;
\STATE When judging using 4 classifier voting rules, whether the number of reliable samples is greater than half of the total sample number of the unseen class, if yes, use 4 classifier voting rules, otherwise, use 3 classifier voting rules to assign pseudo labels to $D^{U}$;
\STATE Generate the pseudo-labeled data set $D_{\text {pseudo }}$, update the training set $D_{\text {train }} \leftarrow D^{S} \cup D_{\text {pseudo }}$;
\STATE Train two LGFAA networks with different structures on $D_{\text {train }}$ to form a collaborative network, and get the pseudo-labels of the two networks to predict $D^{U}$;
\end{algorithmic}
\textbf{Until}
\begin{algorithmic}
\STATE Execute to the fourth iteration, change the voting rule to only use 3 classifier voting rules, and execute it twice
\end{algorithmic}
\end{algorithm}

\section{Experiment}
\subsection{Experimental dataset}

In order to prove the effectiveness of the proposed method, we provide research in two cases. One is a representative zero-shot dataset, and the other is a cylinder liner surface defect data set in the industrial field.

First, experiments are conducted on three representative attribute vector dataset in the ZSL field, of which two small fine-grained data sets are Caltech-UCSD Birds200-2011(CUB) \cite{wah2011caltech} and Sun attribute Database(SUN) \cite{patterson2012sun}. A medium sized coarse-grained data set is Animals with Attribute 2 (AWA2) \cite{xian2018zero}. Each data set has a human-defined common attribute library, which mainly contains attribute characteristics such as "color, wings, crawling, tail", etc. The defined attribute characteristics include both 0/1 type Boolean values and continuous vectors, as shown in Table \ref{tab:2}. Detailed description of these data sets.

\begin{table*}[htb]
\centering
\begin{center}
\caption{Zero-shot dataset} \label{tab:2}
\begin{tabular}{cccc}
\toprule
\multirow{2}*{Dataset} & Number of  & Number of  & Number of seen class/\\
 ~& samples & attributes & Number of unseen class\\
\midrule
AWA2 & 37322 & 85 & 40/10 \\
CUB & 11788 & 312 & 150/50 \\
SUN & 14340 & 102 & 645/72 \\
\bottomrule
\end{tabular}
\end{center}
\end{table*}

And to perform the ZSL task, the dataset needs to be divided into two disjoint subsets. In order to reflect the effectiveness of the method, we separately perform Standard Split (SS) and Propose Split (PS) on the dataset. PS was proposed by literature\cite{xian2018zero}. Since most zero-shot learning methods use the features extracted by the pre-trained network model trained by ImageNet, and the standard cut of the zero-shot dataset has unseen class that overlap with some of the classes in ImageNet. As a result, the ZSL method has higher accuracy in overlapping classes, so a new Split scheme is proposed—Propose Split.

Next is the cylinder liner defect dataset CLSDD in the industrial field. The dataset is provided by a technology company. It has 412 pictures and 6 types of data (crack, wear, convexity, shrinkage, cavitation, normal). According to the zero-shot learning rule, here use the 4+2 mode for experiments, where 4 types are used as the training set and 2 types are used as the test set. As shown in Table \ref{tab:3}, we use a single defect class and a normal class for testing, a total of 5 groups of test set, the remaining data in each group is used as a training set, and the data set is divided into multiple groups for experiments according to different defects.

\begin{table*}[htb]
\centering
\begin{center}
\caption{Each group of parameters of CLSDD dataset} \label{tab:3}
\begin{tabular}{ccccc}
\hline
\toprule
\multirow{2}*{Group} &\multirow{2}*{Training set}& Number of  & \multirow{2}*{Test set} & Number of\\
 ~ & ~ & samples & ~ & test\\
\hline
Group 1 & Wear,Cavi,Conv,Crac & 287 & Shri,Norm & 125 \\
Group 2 & Shri,Cavi,Conv,Crac & 258 & Wear,Norm & 154 \\
Group 3 & Shri,Wear,Conv,Crac & 287 & Cavi,Norm & 125 \\
Group 4 & Shri,Wear,Cavi,Crac & 249 & Conv,Norm & 163 \\
Group 5 & Shri,Wear,Cavi,Conv & 271 & Crac,Norm & 141 \\
\bottomrule
\end{tabular}
\end{center}
\end{table*}

\subsection{Evaluation Metrics}

In the zero-shot learning task, this paper uses the mainstream average TOP-1 accuracy rate as the evaluation Metrics, and judges the performance of ZSL by averaging the TOP-1 accuracy rate of each category:

\begin{equation}
A c c_{y^{u}}=\frac{1}{\|\varepsilon\|} \sum_{c=1}^{\|\varepsilon\|} A c c_{y_{c}^{u}}
\end{equation}

Where $A C C_{y_{c}}^{u}$ represents the Top-1 accuracy rate of the c-th category in the test set. At this time, the unlabeled samples in the test set are only from the unseen class, and $\varepsilon$ represents the total number of unseen class in the test set.

\subsection{Experimental details}

In the experiment, the zero-shot dataset used artificially defined continuous attributes and normalized. The semantic attributes used in the industrial data set were semantic word vectors extracted by the GloVe model.

The ECT method uses Backbone to use GoogleNet\cite{szegedy2015going} and ResNet101\cite{he2016deep} initialized by ImageNet pre-training, and the image size of the input network is adjusted to the size specified by the network input. In the synthetic virtual feature experiment of unseen class, the ridge regression $\lambda$ is set to 1.0, and the virtual feature part of the unseen class generated by the feature mean and standard deviation uses the random.normal function in the ‘Numpy package’. Use the regression algorithm in the ‘Sklearn package’ to select the classifier. The learning rate parameter of the Lasso classifier is 0.001, the alpha parameter of the ridge regression classifier is set to (0.1, 0.5, 1.0, 5.0, 10.0), and Bayesian uses classification. At the same time, the parameters used by the original LFGAA network have not been changed.

In all experiments, the training uses the Adam optimizer for training in an end-to-end manner, with the parameter $\beta 1$ set to 0.9, $\beta 2$ to 0.999, and the learning rate to 0.001. The input dimension of the semantic embedding module is the image feature dimension output by Backbone. The output dimension is the same as the dimension of the latent attribute space output. The balance parameter of the training semantic embedding module is set to 0.1, the feedback degree parameter of the feedback mechanism is set to 0.01, and the feedback degree parameter is set to 0.005 when using the industrial data set experiment.

\subsection{Zero-shot dataset experiment comparison }

We compare the proposed method with several latest inductive ZSL methods \cite{liu2019attribute,akata2015evaluation,kodirov2017semantic,jiang2018learning,annadani2018preserving,tong2019hierarchical,zhang2020towards}, and transductive ZSL methods \cite{liu2019attribute,zhang2020towards,song2018transductive,ye2019progressive,wan2019transductive,khare2020generative}, where the inductive method means that the model can only use the labeled training set in the training phase, while the transductive method can use both the labeled training set and the unlabeled test set in the training phase.

The experimental results are shown in Table \ref{tab:4}, where 'I' represents the inductive zero-shot image classification, 'T' represents the transductive zero-shot image classification, and '—' indicates that this result is not published in the literature, black means the best. For the inductive zero-shot image classification method, the experimental results of the SF-LFGAA method are slightly weaker in the SUN dataset under PS, but the rest are better than other embedding-based methods. Compared with the baseline method LGFAA, it has been improved, especially the CUB dataset, which has been improved by 4.7$\%$ and 5.2$\%$ respectively. At the same time, LFGAA proposed its own transductive method SA. After combining it with our proposed SF-LFGAA, except that the AWA2 data set under PS is slightly weaker, others have also been improved, especially the CUB dataset, which is compared with the original baseline method has been improved by 3.4$\%$ and 4.5$\%$. It can be seen from the experimental results that after the semantic embedding module and feedback mechanism are added, the latent attributes of the adjustment are more discriminative. Since the latent attributes also affect the attribute attention and visual features, it further reduces the ambiguity between target categories.

\begin{table*}[htb]
\centering
\begin{center}
\caption{Comparisons in the ZSL dataset} \label{tab:4}
\begin{tabular}{c|ccccccc}
\toprule
\multirow{2}*{~} & \multirow{2}*{Method} & \multicolumn{2}{c}{AWA2} & \multicolumn{2}{c}{CUB} & \multicolumn{2}{c}{SUN} \\
~&~& SS & PS & SS & PS & SS & PS \\
\hline
\multirow{8}*{I} & SJE\cite{akata2015evaluation} & 69.5 & 61.9 & 55.3 & 53.9 & 57.1 & 53.7 \\
~ & SAE\cite{kodirov2017semantic} & 80.7 & 54.1 & 33.4 & 33.3 & 42.4 & 40.4 \\
~ & CDL\cite{jiang2018learning} & 79.5 & 67.9 & 54.5 & 54.5 & 61.3 & 63.6 \\
~ & PSR-ZSL\cite{annadani2018preserving} & — & 63.8 & — & 56.0 & — & 61.4 \\
~ & DLFZRL\cite{tong2019hierarchical} & — & 63.7 & — & 57.8 & — & 59.3 \\
~ & TEDE\cite{zhang2020towards} & — & 66.5 & — & 57.1 & — & 62.4 \\
~ & LFGAA\cite{liu2019attribute} & 84.3 & 68.1 & 67.6 & 67.6 & 62.0 & 61.5 \\
~ & SF-LFGAA(our) & 85.7 & 69.3 & 72.3 & 72.8 & 62.6 & 62.0 \\
\hline
\multirow{9}*{T} & QFSL\cite{song2018transductive} & 84.8 & 79.3 & 69.7 & 72.1 & 61.7 & 58.3 \\
~ & PREN\cite{ye2019progressive} & 96.1 & 78.6 & 66.6 & 66.4 & 63.2 & 62.8 \\
~ & VCAM\cite{wan2019transductive} & 93.9 & 78.2 & 74.2 & 71.7 & — & — \\
~ & Trans-ADA\cite{khare2020generative} & — & 78.6 & — & 74.2 & — & 65.5 \\
~ & Trans-TEDE\cite{zhang2020towards} & — & 77.5 & — & 67.8 & — & 61.6 \\
~ & LFGAA + SA\cite{liu2019attribute} & 94.4 & 84.8 & 79.7 & 78.9 & 64.0 & 66.2 \\
~ & SF-LFGAA+ SA(our) & 95.3 & 83.1 & 83.1 & 83.4 & 64.7 & 66.5 \\
~ & LFGAA + ECT(our) & 94.7 & 82.1 & 81.3 & 81.7 & 69.8 & 70.1 \\
~ & SF-LFGAA + ECT(our) & 96.3 & 83.9 & 83.7 & 83.9 & 70.1 & 70.5 \\
\bottomrule
\end{tabular}
\end{center}
\end{table*}

We also proposed our own transductive method ECT, and combined LFGAA and SF-LFGAA with ECT. Compared with the baseline methods LFGAA and SF-LFGAA that add SA, it can be seen that all items exceed the original benchmark method, indicating that our proposed ECT method can alleviate the domain shift problem to a certain extent.

\subsection{Industrial dataset experiment comparison}

By consulting relevant information, we found that there has never been an experiment in the past without sample-based model training to achieve defect detection and classification of cylinder liners in the industrial field.

Therefore, we focused on the alternative methods LGFAA and LGFAA + SA. The results under our improved method are shown in Table \ref{tab:5}. When LFGAA is compared with SF-LFGAA method, in groups 1, 3, and 5, regardless of whether it is combined with the transductive method SA, except for group 4, almost all other groups have been significantly improved. Compared with the original network LFGAA, the performance of the SF-LFGAA of the group 4 dataset is lower. It is speculated that there is still a certain gap between the word vector features provided by the GloVe model and the attributes artificially defined by experts in the zero-shot dataset. which causes the effect of some grouping experiments to decrease. At the same time, after we combine the LFGAA and SF-LFGAA methods with the transductive method ECT, the performance of the model has been significantly improved. Compare LFGAA + ECT with LFGAA + SA, it is only slightly weaker in the experiment of group 5, but it has been significantly improved in other groups. After SF-LFGAA+ and ECT are combined, they all exceed the baseline method. To sum up, we have proved the effectiveness of the SF-LFGAA and ECT methods through experiments

\begin{table*}[htb]
\centering
\begin{center}
\caption{Comparisons in the CLSDD dataset} \label{tab:5}
\begin{tabular}{cccccc}
\hline
\toprule
Methond&Group1&Group2&Group3&Group4&Group5\\
\hline
LFGAA & 72.6 & 53.7 & 68.5 & 68.9 & 62.2 \\
SF-LFGAA(our) & 79.6 & 61.0 & 74.0 & 58.4 & 73.0 \\
LFGAA + SA & 87.4 & 70.0 & 86.3 & 74.9 & 79.9 \\
SF-LFGAA + SA(our) & 92.3 & 76.4 & 93.4 & 73.9 & 81.9 \\
LFGAA + ECT(our) & 91.7 & 85.7 & 95.4 & 82.7 & 78.3 \\
SF-LFGAA + ECT(our) & 93.6 & 87.7 & 94.4 & 87.7 & 83.0 \\
\bottomrule
\end{tabular}
\end{center}
\end{table*}

\section{Analysis of experiment modules}
\subsection{Effective semantic embedding module}
Although the output of the LFGAA network itself has a semantic vector, when the image is predicted, the vector output from the semantic space will be combined with the attribute attention as the predictive attribute of the image. Due to the attribute attention mechanism, the non-representative attributes do not have high activation, so when the processed semantic vector is used to adjust the latent attributes, the model performance may not be significantly improved. At the same time, we also consider the situation that the semantic vector output by the LFGAA network is not combined with the attribute attention, and use it as the input of the feedback mechanism

\begin{figure}[t]
\centering
\includegraphics[scale=0.26]{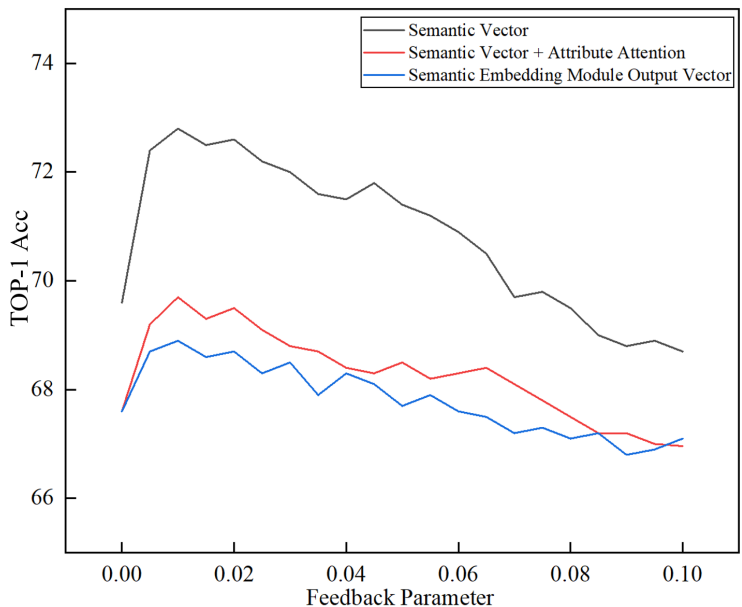}
\caption{Model performance comparison of different semantic vectors}
\label{fig:8}
\end{figure}

Taking the CUB data set under PS as an example, the following experiments were carried out. The semantic vector output by the LGFAA network, the semantic vector combined with the attribute attention and the vector output by the semantic embedding module are respectively used as the input of the feedback mechanism. The results shown in Figure.\ref{fig:8} are obtained under different feedback degree parameters.

It can be seen from the Figure. 8 that when the feedback degree parameter is 0.01, the model performance reaches the highest. After the semantic embedding module is added, even if it is not feedback (that is, the feedback parameter is 0), the performance of the model is improved by about 2$\%$. At the same time, we found that under appropriate feedback parameters, the semantic vector and the semantic vector combined with the attribute attention, although also played a role in improving the performance of the model, but the output vector of the semantic embedding module is used as the input of the feedback mechanism is more effective for the model, reflecting the importance of semantic embedding module.

\subsection{Effective feedback mechanism}

The purpose of constructing the feedback mechanism is to adjust the output of the semantic embedding module to the latent attributes, making the latent attributes more discriminative. In order to demonstrate the effectiveness of the feedback mechanism, we take the PS of the CUB dataset as an example, as shown in Table \ref{tab:6}. The top-1 accuracy of some categories of the CUB dataset under the baseline method LGFAA and the improved method SF-LFGAA. It can be seen that the improved method has significantly improved accuracy under both inductive and transductive zero-shot image classification. (Because of the excessive number of unseen classes in the CUB dataset, it is difficult to present them in the later visualization, so only some of the classes are shown).

\begin{table*}[htb]
\centering
\begin{center}
\caption{Feedback mechanism before and after the addition of comparative experiments} \label{tab:6}
\begin{tabular}{ccccc}
\hline
\toprule
Unseen class&LFGAA&SF-LFGAA&LFGAA+SA&SF-LFGAA+SA\\
\hline
Bronzed&\multirow{2}*{68.3} &\multirow{2}*{86.7}&\multirow{2}*{85.0}&\multirow{2}*{90.0}\\
Cowbird&~&~&~&~\\
Blue Winged&\multirow{2}*{58.3}&\multirow{2}*{88.3}&\multirow{2}*{76.7}&\multirow{2}*{85.0}\\
Warbler&~&~&~&~\\
Kentucky&\multirow{2}*{50.8}&\multirow{2}*{64.4}&\multirow{2}*{91.5}&\multirow{2}*{93.2}\\
Warbler&~&~&~&~\\
Common&\multirow{2}*{46.7}&\multirow{2}*{61.7}&\multirow{2}*{66.7}&\multirow{2}*{93.3}\\
Yellowthroat&~&~&~&~\\
\bottomrule
\end{tabular}
\end{center}
\end{table*}

\begin{figure}[t]
\centering
\includegraphics[scale=0.3]{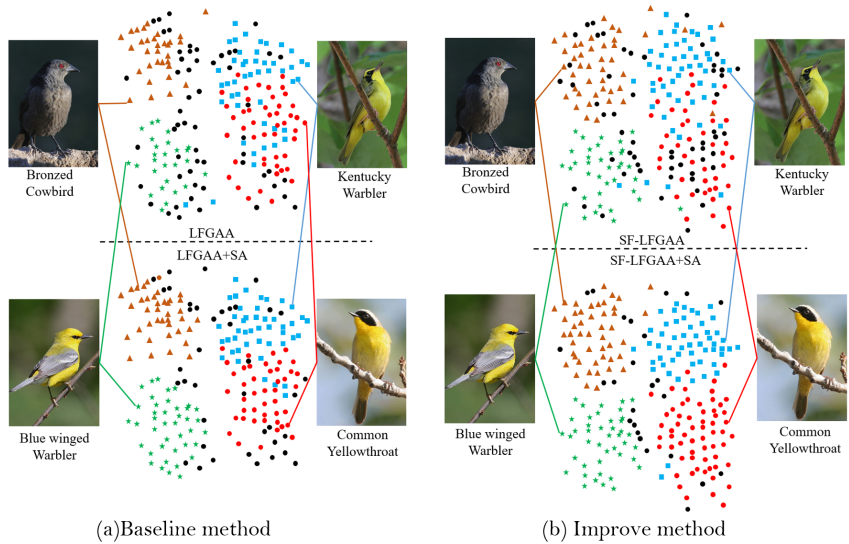}
\caption{Visualization of visual feature distribution of different training methods}
\label{fig:9}
\end{figure}

At the same time, we use T-SNE to visualize the prediction distributions of these four categories. Figure.\ref{fig:9}(a) and Figure.\ref{fig:9}(b) represent the baseline method and the improvement method, respectively. The "black circles" represent other categories predicted by the network. Since some of the predicted categories are distributed far away in space, they mainly show the predicted distributions around the four categories. It can be seen from Figure.\ref{fig:9}(a) that the two methods, LGFAA and LGFAA+SA, share the same visual feature network, resulting in the same feature distribution, which differs only in the prediction results. While the feedback mechanism in the method in Figure.\ref{fig:9}(b) modulates the latent attributes, it also affects the visual features, so the feature distribution is different from that in Fig.\ref{fig:9}(a). The Figure.\ref{fig:9} also reflects the fact that there are some mixed categories in the prediction distribution of LFGAA, SF-LFGAA improves this situation and brings good classification performance.

\subsection{Analysis of Ensemble Co-training algorithm}

In this section, we directly learn the mapping of visual features to semantics (the loss function uses the BCE loss function) to study the influence of the labeling rules in the algorithm on the overall model performance, and explain why the attribute classifier is chosen so.
\paragraph{Semantic classifier} This paper uses a number of different semantic classifiers to complete the prediction of image semantic attributes. The overall performance of the network model depends on the performance of the classifier, and the performance of the classifier depends on the image features and semantic attributes of the dataset. In the aspect of image features, ResNet101 is used for extraction, and the semantic attributes used in the experiment use the continuous attributes of the data set itself. Because the network itself is trained through an end-to-end neural network, we use the CUB data set under PS and only compare the current mainstream regression algorithms.

The experimental results are shown in Table \ref{tab:7}, from which it can be seen that Lasso regression, ridge regression, Bayesian ridge regression, Support Vector Machine and Linear regression show high accuracy. However, the Support Vector Machine takes a long time. The reason why Linear regression is rejected is that the ridge regression classifier can set multiple parameter values for cross-validation to select the best prediction result. This also causes the prediction result to have a strong relationship with linear regression. Similarity, and has more advantages than linear regression in terms of performance. We improve the generalization performance by setting multiple classifiers to reduce the risk of a single classifier entering the local minimum point, and finally use Lasso regression, ridge regression, and Bayes ridge regression as semantic classifiers.

\begin{table*}[htb]
\centering
\begin{center}
\caption{Comparison of zero-sample classification results and training time of different classifiers} \label{tab:7}
\begin{tabular}{ccc}
\hline
\toprule
Classifier&Accuracy$(\%)$&Time(s)\\
\hline
LASSO regression & 46.5 & 135 \\
Ridge regression & 48.7 & 17 \\
Bayesian Ridge Regression & 47 & 980 \\
Support Vector Regression & 44.5 & 1351 \\
Linear regression & 47.5 & 10 \\
KNN regression & 31.2 & 31 \\
Random forest & 29.7 & 6867 \\
Bagging regression & 25.7 & 120 \\
\bottomrule
\end{tabular}
\end{center}
\end{table*}

\subsubsection{Marking rules}

\paragraph{Marking rules} In the paper, we use the voting mechanism in ensemble learning to judge whether the selected sample is reliable. The voting mechanism refers to combining multiple classifiers, and then according to the results of each classifier's prediction, execute the minority to obey the majority, and use the majority prediction result as the final result. Similarly, in the pseudo-label prediction phase, whether the labeled pseudo-label is reliable depends on the number of votes Z (That is, the number of different classifiers predicting the same result) of the predicted label. This paper formulates its own pseudo-label marking rules by observing the experimental results. The pseudo-label marking rules are described in detail in ECT Algorithm.

This paper uses the SUN dataset under the PS to conduct experiments. In the experiment, we found that the performance of the model depends on the number and accuracy of unlabeled samples by setting a different number of votes. Table \ref{tab:8}shows that under the different number of votes of the classifier and the marking rules by ourselves, the first iteration gives the number of unknown pseudo-labels, the accuracy of labeled data, and the accuracy of the network model trained with labeled data. It can be seen that when the number of votes is in the median range, the accuracy rate reaches a higher level. Since the number of votes is 2, 3, and 4, almost all labels are assigned, so it is difficult to greatly improve the performance in the later stage.

\begin{table*}[htb]
\centering
\begin{center}
\caption{Unseen class sample label number, pseudo label accuracy and model accuracy under different voting numbers} \label{tab:8}
\begin{tabular}{cccc}
\hline
\toprule
Number of& Number of  & Label  & Model accuracy after training\\
votes& marks & accuracy & with pseudo-labels\\
\hline
2 & 1492 & 58.3 & 57.3 \\
3 & 1480 & 58.5 & 57.5 \\
4 & 1337 & 62.3 & 58.0 \\
5 & 1096 & 66.8 & 57.9 \\
6 & 832 & 73.4 & 57.6 \\
7 & 629 & 78.3 & 55.8 \\
8 & 369 & 85.1 & 53.7 \\
Our & 852 & 74.3 & 57.8 \\
\bottomrule
\end{tabular}
\end{center}
\end{table*}

\begin{figure}[t]
\centering
\includegraphics[scale=0.4]{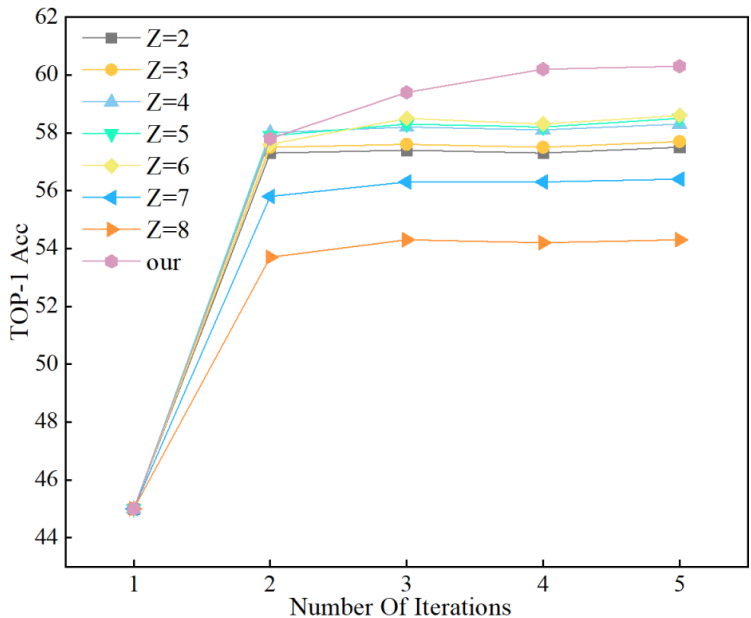}
\caption{Comparison of the number of different votes}
\label{fig:10}
\end{figure}

This paper also tested the accuracy changes of the zero-shot classification model under different voting numbers and the marking rules we formulated. It can be seen from Figure.\ref{fig:10}, in the third cycle, the accuracy rate is no longer increasing. We believe that changing the number of votes can increase this trend, so we have formulated a voting rule to change the voting rule to only use three classifiers when the loop is executed for the fourth time. By observing the experimental results of different voting numbers in Table \ref{tab:8} and Figure.\ref{fig:10}, we have formulated our own marking rules. It can be seen that our marking rules have achieved the best performance. It also reflects that in the later cycle, the model performance does not change significantly when the number of marked samples and the accuracy rate do not change.

\section{Conclusion}

In this paper, considering the difficulty of sample collection in the industrial field, we apply the zero-shot learning technology to the industrial field. As far as we know, this is the first attempt to detect and classify cylinder liner defects without target samples.

We propose a semantic feedback-based LGFAA network to alleviate the problem that the output latent attributes and artificially defined attributes in the zero-shot classification network LFGAA are different in the semantic space, which causes the model performance to decrease. At the same time, it is aimed at the zero-shot classification. For the domain shift problem that exists in the task, an Ensemble co-training algorithm is proposed to adaptively reduce the prediction error in image tag embedding from multiple angles. We prove the effectiveness of the proposed method through relevant comparative experiments in different zero-shot image classification dataset, and at the same time extend the method to the cylinder liner surface defect dataset in the industrial field, which proves the feasibility of the proposed method in this field.

\bibliography{mybibfile}

\end{document}